\documentclass{article}
\usepackage{spconf}
\ninept

\usepackage[utf8]{inputenc} 
\usepackage[T1]{fontenc}    
\usepackage{hyperref}       
\usepackage{url}            
\usepackage{booktabs}       
\usepackage{amsfonts}       
\usepackage{nicefrac}       
\usepackage{microtype}      
\usepackage{siunitx}
\usepackage{enumerate}
\usepackage{float}

\hypersetup{allcolors=blue,colorlinks=true}

\usepackage{graphicx}
\usepackage{amsmath}
\usepackage{amsthm}
\usepackage{amssymb}
\usepackage{amsfonts}
\usepackage{mathrsfs}
\usepackage{multirow}
\usepackage{algorithm} 
\usepackage{subfigure} 
\usepackage{thm-restate}

\usepackage{bm}
\newcommand{\R}{\mathbb{R}}
\newcommand{\B}{\mathbb{B}}
\newcommand{\bA}{\bm{A}}
\newcommand{\bE}{\bm{E}}
\newcommand{\bI}{\bm{I}}
\newcommand{\bS}{\bm{S}}
\newcommand{\bQ}{\bm{Q}}
\newcommand{\bR}{\bm{R}}
\newcommand{\bU}{\bm{U}}
\newcommand{\bV}{\bm{V}}
\newcommand{\bX}{\bm{X}}
\newcommand{\bOmega}{\bm{\Omega}}
\newcommand{\bu}{\bm{u}}
\newcommand{\bv}{\bm{v}}
\newcommand{\bx}{\bm{x}}
\newcommand{\by}{\bm{y}}
\newcommand{\bz}{\bm{z}}

\usepackage{bbm}

\newcommand{\bzero}{\mathbf{0}}

\DeclareMathOperator*{\argmin}{arg\,min}
\DeclareMathOperator*{\argmax}{arg\,max}
\DeclareMathOperator{\prox}{\textsf{prox}}
\DeclareMathOperator{\proj}{\textsf{proj}}
\DeclareMathOperator{\chol}{\textsf{chol}}
\DeclareMathOperator{\Tr}{Tr}

\newtheorem{assumption}{Assumption}
\newtheorem{lemma}{Lemma}

\usepackage[natbib=true,
            bibstyle=ieee,
            citestyle=ieee,
            uniquename=false,
            uniquelist=false,
            maxcitenames=2,
            mincitenames=1,
            mincrossrefs=99]{biblatex}
\renewbibmacro{in:}{\ifentrytype{article}{}{\printtext{\bibstring{in}\intitlepunct}}}
\DefineBibliographyStrings{english}{url = Available at}

\title{Sparse and Functional Principal Components Analysis}

\twoauthors{Genevera I. Allen\thanks{GA is also affiliated with the Jan and Dan Duncan Neurological Research Institute at the Baylor College of Medicine, Houston, TX 77030. GA acknowledges support from NSF DMS-1554821, NSF NeuroNex-1707400, and NSF DMS-126405. MW acknowledges support from the NSF Graduate Research Fellowship Program under grant number 1450681. The authors thank Dr. Yue Hu for assistance preparing the EEG Data used in Section \ref{sec:eeg} and Luofeng Liao for useful discussions.}}{%
            Departments of Statistics, CS, and ECE\\
            Rice University\\
            Houston, TX 77005\\
            \href{mailto:gallen@rice.edu}{gallen@rice.edu}}{%
            Michael Weylandt}{%
            Department of Statistics\\
            Rice University\\
            Houston, TX 77005\\
            \href{mailto:michael.weylandt@rice.edu}{michael.weylandt@rice.edu}}

\begin{document}
\begin{refsection}
\maketitle

\begin{abstract}
Regularized variants of Principal Components Analysis, especially Sparse PCA and Functional PCA, are among the most useful tools for the analysis of complex high-dimensional data. Many examples of massive data have both sparse and functional (smooth) aspects and may benefit from a regularization scheme that can capture both forms of structure. For example, in neuro-imaging data, the brain's response to a stimulus may be restricted to a discrete region of activation (spatial sparsity), while exhibiting a smooth response within that region. We propose a unified approach to regularized PCA which can induce both sparsity and smoothness in both the row and column principal components. Our framework generalizes much of the previous literature, with sparse, functional, two-way sparse, and two-way functional PCA all being special cases of our approach.  Our method permits flexible combinations of sparsity and smoothness that lead to improvements in feature selection and signal recovery, as well as more interpretable PCA factors. We demonstrate the efficacy of our method on simulated data and a neuroimaging example on EEG data.  
\end{abstract}

\section{Introduction}

Principal Component Analysis (PCA) is a fundamental technique for dimension reduction, pattern recognition, and visualization of multivariate data. In the early 2000s, researchers noted that naive extensions of PCA to the high-dimensional setting produced unsatisfactory results, a finding later confirmed by advances in random matrix theory \citep{Johnstone:2009}. To address this limitation, many regularized variants of PCA were proposed, wherein the principal components were estimated under smoothness or sparsity assumptions \citep{Silverman:1996,Huang:2008,Huang:2009,Witten:2009,Allen:2011,Allen:2014}. Rather than reviewing this large literature, we instead refer the reader to the recent reviews of \citet{Hall:2011}, focusing on functional (smooth) PCA (FPCA) and of \citet{Zou:2018}, focusing on sparse PCA (SPCA).

Given the importance of both FPCA and SPCA, it is natural to ask whether it is possible to combine these approaches, yielding a unified approach to \emph{sparse and functional PCA} (SFPCA). We show that this is indeed possible and present a unified optimization framework for doing so. Our proposed approach unifies much of the existing literature on regularized PCA; standard PCA, SPCA, FPCA, two-way SPCA, and two-way FPCA are all special cases of our approach, suggesting that it is, in some sense, the ``correct'' generalization.

Our unified SFPCA method enjoys many advantages over existing approaches to regularized PCA: i) because it allows for arbitrary degrees and forms of regularization, it is conducive to data-driven determination of the appropriate types and amount of regularization for a given problem; ii) because it unifies many existing methods, it inherits the desirable properties of both SPCA and FPCA, including superior signal recovery, automatic feature selection, and improved interpretability; and iii) it admits a tractable, efficient, and theoretically well-grounded algorithm.  

Throughout this paper, we adopt the low-rank perspective on PCA and assume that our observed data $\bX \in \R^{n \times p}$ arises from a low-rank structure $\bX = \sum_{k = 1}^K d_k \bu_k\bv_k^T + \bE$, where the elements of $\bE$ are independently and identically distributed with mean 0. We refer to the vectors $\{\bu_k\}_{k = 1}^K \in \R^n$ and $\{\bv_k\}_{k = 1}^K \in \R^p$ as the left and right singular vectors respectively. Given $\bX$, its leading singular vectors can be estimated by solving the singular value problem: \begin{equation}\argmax_{\bu \in \overline{\B}^n, \bv \in \overline{\B}^p} \bu^T\bX\bv \label{eqn:svp} \end{equation} where $\overline{\B}^n = \{\bx \in \R^n: \|\bx\|_2 \leq 1\}$ is the unit ball in $\R^n$. (Some authors require $\|\bu\|_2 = \|\bv\|_2 = 1$, but, because the objective is linear in both $\bu$ and $\bv$, solutions to \eqref{eqn:svp} lie on the boundary and this does not fundamentally change the problem.) Since the following singular vectors can be recovered by solving Problem \eqref{eqn:svp} on a ``deflated'' $\bX$, throughout this paper we principally focus on the leading singular vectors. Assuming that $\bX$ has previously been centered, this approach is known to be equivalent to applying the eigenproblem formulation of PCA to both $\bX\bX^T$ and $\bX^T\bX$.

\section{A Sparse and Functional Singular Value Formulation of PCA}
Taking the singular value problem \eqref{eqn:svp} as a starting point, \citet{Huang:2009} proposed two-way FPCA by adding a product smoothness penalty \[\argmax_{\bu \in \overline{\B}^n, \bv \in \overline{\B}^p} \bu^T\bX\bv - \lambda\|\bu\|_{\bS_{\bu}}^2\|\bv\|_{\bS_{\bv}}^2\] where $\|\bu\|_{\bS_{\bu}}^2 = \bu^T\bS_{\bu}\bu$ for some positive-definite $\bS_{\bu}$ (similarly for $\bv$). Typically, we take $\bS_{\bu} = \bI + \alpha_{\bu} \bOmega_{\bu}$ where $\bOmega_{\bu}$ is the second- or fourth-difference matrix, so that the $\|\bu\|_{\bS_{\bu}}^2$ penalty term encourages smoothness in the estimated singular vectors. Similarly, \citet{Allen:2014} proposed two-way SPCA by adding sparsity inducing penalties to the singular value problem \eqref{eqn:svp}: \[\argmax_{\bu \in \overline{\B}^n, \bv \in \overline{\B}^p} \bu^T\bX\bv - \lambda_{\bu}P_{\bu}(\bu) - \lambda_{\bv}P_{\bv}(\bv)\] where $P_{\bu}$ and $P_{\bv}$ are sparsity inducing penalties. (This is the Lagrangian form of the method of \citet{Witten:2009}.) Given the success of these two methods, it is perhaps natural to perform SFPCA by adding both smoothness and sparsity penalties to Problem \eqref{eqn:svp}: \begin{equation} \argmax_{\bu \in \overline{\B}^n, \bv \in \overline{\B}^p} \bu^T\bX\bv - \lambda_{\bu}P_{\bu}(\bu) - \lambda_{\bv}P_{\bv}(\bv) - \lambda\|\bu\|_{\bS_{\bu}}^2\|\bv\|_{\bS_{\bv}}^2 \label{eqn:bad_sfpca} \end{equation}
Surprisingly, this natural generalization fails, often spectacularly!

\begin{figure}
\centering
\includegraphics[width=1.5in]{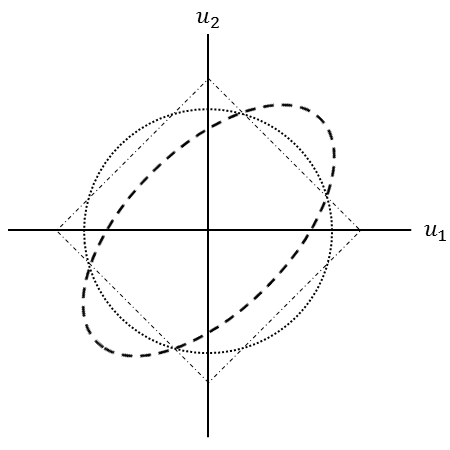}
\caption{Three constraints implicit in the ill-posed naive formulation of SFPCA: sparsity constraint ($\ell_1$-ball), unit norm ($\ell_2$-ball), and smoothness (elliptical region). In general, it is difficult for a point to lie on the boundary of all three regions simultaneously, leading to degenerate solutions to Problem \eqref{eqn:bad_sfpca}.}
\label{fig:balls}
\end{figure}

To see why this occurs, we note that Problem \eqref{eqn:bad_sfpca}, with $\bv$ held fixed, is actually attempting to satisfy three different constraints on $\bu$ independently: a standard norm constraint, a smoothness constraint, and a sparsity constraint. As shown in Figure \ref{fig:balls}, unless all three regularization parameters ($\lambda, \alpha_{\bu}, \lambda_{\bu}$) are carefully chosen, this results in a form of ``regularization masking,'' whereby it is impossible for the solution to Problem \eqref{eqn:bad_sfpca} to satisfy all constraints simultaneously. For the general case of two-way SFPCA, where we impose multiple constraints on both $\bu$ and $\bv$, this phenomenon is compounded. 

To address the problem of regularization masking, we instead propose the following formulation of SFPCA: 
\begin{equation} 
\argmax_{\bu \in \overline{\B}^n_{\bS_{\bu}}, \bv \in \overline{\B}^p_{\bS_{\bv}}} \bu^T\bX\bv - \lambda_{\bu} P_{\bu}(\bu) - \lambda_{\bv} P_{\bv}(\bv) \label{eqn:sfpca}
\end{equation}
where $\overline{\B}^n_{\bS_{\bu}}$ is the unit ellipse of the $\bS_{\bu}$-norm, \emph{i.e.}, $\overline{\B}^n_{\bS_{\bu}} = \{\bu \in \R^n: \bu^T\bS_{\bu}\bu \leq 1\}$. As we will see below, this formulation is the ``correct'' generalization of many of the regularized PCA formulations previously proposed in the literature. Comparing our SFPCA formulation \eqref{eqn:sfpca} with the naive formulation \eqref{eqn:bad_sfpca}, we note two key differences: firstly, we only use a sparsity penalty in the objective function, moving the smoothness terms to the constraints to avoid regularization masking; secondly, we replace the unit ball constraint with a more general unit ellipse constraint. Since the unit ball constraint exists only to ensure identifiability of Problem \eqref{eqn:svp}, replacing it with a unit ellipse constraint simplifies the problem and ameliorates regularization masking. 
The benefits of this reformulation in eliminating regularization masking are formalized in Theorem \ref{thm:kkt} below. 

Before proceeding, we make two regularity assumptions which we will use throughout our subsequent theoretical analysis: 
\begin{assumption}
In the SFPCA problem \eqref{eqn:sfpca}, with $\bS_{\bu} = \bI + \alpha_{\bu}\bOmega_{\bu}$ and $\bS_{\bv} = \bI + \alpha_{\bv}\bOmega_{\bv}$ for $\alpha_{\bu}, \alpha_{\bv} \geq 0$, the following hold: \label{ass:general}
\begin{enumerate}[(i)]
  \item The smoothing matrices $\bOmega_{\bu}, \bOmega_{\bv}$ are positive semi-definite. 
  \item The penalty terms $P_{\bu}, P_{\bv}$ take values in $\R_{\geq 0}$ and are positive homogeneous of order one, \emph{i.e.}, $P(c\bx) = cP(\bx)$ for all $c > 0$ and all $\bx$. 
\end{enumerate}
\end{assumption}
\noindent Under these assumptions, our formulation of SFPCA \eqref{eqn:sfpca} is well-posed and avoids many of the pathologies associated with other formulations:
\begin{restatable}{theorem}{kkt} \label{thm:kkt}
Suppose Assumption \ref{ass:general} holds and let $(\bu^{*}, \bv^{*})$ be the optimal points of the SFPCA problem \eqref{eqn:sfpca}. Then the following hold:
\begin{enumerate}[(i)]
  \item There exist values $\lambda_{\bu}^{\max}$ and $\lambda_{\bv}^{\max}$ such that, if $\lambda_{\bu} \geq \lambda_{\bu}^{\max}$ or if $\lambda_{\bv} \geq \lambda_{\bv}^{\max}$, then the solution to Problem \eqref{eqn:sfpca} is trivial in the sense $(\bu^*, \bv^*) = (\bzero, \bzero)$.
  \item If $\lambda_{\bu} < \lambda_{\bu}^{\max}$ and $\lambda_{\bv} < \lambda_{\bv}^{\max}$, the SFPCA solution $(\bu^{*}, \bv^{*})$ depends on all (non-zero) regularization parameters.  
  \item $\|\bu^{*}\|_{\bS_{\bu}}$ is equal to either $1$ or $0$, with the latter occurring only when $\lambda_{\bu} \geq \lambda_{\bu}^{\max}$ or $\lambda_{\bv} \geq \lambda_{\bv}^{\max}$. An analogous result holds for $\bv^{*}$.
  \item $(\bu^{*}, \bv^{*})$ do not suffer from scale non-identifiability. (That is, $(c\bu^*, c^{-1}\bv^*)$ is not a solution for any $c \geq 0$ except $c = 1$.)
\end{enumerate}
\end{restatable}
\noindent The requirements of Assumption \ref{ass:general} are in fact quite weak and allow for nearly all the sparsity and smoothness structures previously proposed in the literature, including convex sparsity-inducing penalties (\emph{e.g.}, the lasso \citep{Tibshirani:1996}), structured-sparsity penalties such as the group or fused lasso \citep{Tibshirani:2005,Yuan:2006}, and penalties based on the generalized lasso \citep{Tibshirani:2011}, as well as more exotic penalties such as the \textsc{slope} penalty of \citet{Bogdan:2015}. As the following theorem shows, for various choices of the regularization parameters, SFPCA can yield the solution to standard PCA (SVD), SPCA, FPCA, two-way SPCA, and two-way FPCA: 
\begin{restatable}{theorem}{equivalence} \label{thm:equivalence}
Suppose Assumption \ref{ass:general} holds and let $(\bu^{*}, \bv^{*})$ be the optimal points of the SFPCA problem \eqref{eqn:sfpca}. Then the following hold (up to a sign factor and unit scaling): 
\begin{enumerate}
  \item[(i)] If $\lambda_{\bu}, \lambda_{\bv}, \alpha_{\bu}, \alpha_{\bv} = 0$, then $\bu^{*}$ and $\bv^{*}$ are the first left and right singular vectors of $\bX$.
  \item[(ii)] If $\lambda_{\bu}, \alpha_{\bu}, \alpha_{\bv} = 0$, then $\bu^{*}$ and $\bv^{*}$ are equivalent to the SPCA solution of \citet{Shen:2008}.
  \item[(iii)] If $\alpha_{\bu}, \alpha_{\bv} = 0$, then $\bu^{*}$ and $\bv^{*}$ are equivalent to the two-way SPCA solution in~\citet{Allen:2014}, itself a special case of two-way sparse GPCA with the generalizing operators $\bQ, \bR$ both identity matrices. (This is also the Lagrangian form of \citet{Witten:2009}.)
  \item[(iv)] If $\lambda_{\bu}, \lambda_{\bv}, \alpha_{\bu} = 0$, then $\bu^{*}$ and $\bv^{*}$ are equivalent to the FPCA solution of \citet{Silverman:1996} and \citet{Huang:2008}.
  \item[(v)] If $\lambda_{\bu}, \lambda_{\bv} = 0$, then $\bu^{*}$ and $\bv^{*}$ are equivalent   to the two-way FPCA solution of \citet{Huang:2009}.
\end{enumerate}
For parts (ii) and (iii), equivalencies hold for the appropriate $P_{\bu}(\cdot)$ and $P_{\bv}(\cdot)$ employed in the referenced papers.
\end{restatable} 



\begin{figure*}[t]
\centering
\vspace{-0.6in}
\includegraphics[height=2in,width=6in]{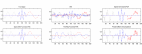}
\vspace{-0.4in}
\caption{Simulated factors used for the simulation study in Section \ref{sec:simulation} and estimates thereof: $\bv_{1}$ (red, dotted-dashed); $\bv_{2}$ (blue, dashed); and $\bv_{3}$ (black, dotted). Only SFPCA is able to simultaneously identify the spatial sparsity and smooth structure of the sinusoidal pulses.}
\label{fig:simulationI}
\end{figure*}

\section{Computation of Sparse and Functional Principal Components}
We next present an efficient algorithm for computing sparse and functional components by solving Problem \eqref{eqn:sfpca}. The key to our algorithm is the observation that, if $P_{\bu}, P_{\bv}$ are convex functions, then Problem \eqref{eqn:sfpca} is a bi-concave problem in $\bu$ and in $\bv$, where each subproblem is equivalent to a penalized regression problem. This suggests an alternating proximal gradient ascent strategy, which yields the following rank-one SFPCA Algorithm, where $\lambda_{\max}(\bA)$ is the leading eigenvalue of $\bA$ and $\prox_{f(\cdot)}(\bz) = \argmin_{\bx} \frac{1}{2}\|\bx - \bz\|_2^2 + f(\bx)$ is the proximal operator of $f$:
\begin{algorithm}[H] 
\caption{Rank-1 SFPCA Algorithm (Proximal Gradient Variant)}\label{alg:sfpca}
\begin{enumerate}
  \item Initialize $\hat{\bu}, \hat{\bv}$ to the leading singular vectors of $\bX$ and set $L_{\bu} = \lambda_{\max}(\bS_{\bu})$ and $L_{\bv} = \lambda_{\max}(\bS_{\bv})$
  \item Repeat until convergence:
  \begin{enumerate}[(a)]
    \item $\bu$-subproblem: repeat until convergence: 
    \begin{align*}
    \bu &:= \prox_{\frac{\lambda_{\bu}}{L_{\bu}} P_{\bu}(\cdot)}\left(\bu + L_{\bu}^{-1}\left(\bX\hat{\bv} - \bS_{\bu}\bu\right)\right) \\
    \hat{\bu} &:= \begin{cases} \bu & \|\bu\|_{\bS_{\bu}} \leq 1 \\ \bu / \|\bu\|_{\bS_{\bu}} & \text{ otherwise} \end{cases}
    \end{align*}
    \item $\bv$-subproblem: repeat until convergence: 
    \begin{align*}
    \bv &:= \prox_{\frac{\lambda_{\bv}}{L_{\bv}} P_{\bv}(\cdot)}\left(\bv + L_{\bv}^{-1}\left(\bX^T\hat{\bu} - \bS_{\bv}\bv\right)\right) \\
    \hat{\bv} &:= \begin{cases} \bv & \|\bv\|_{\bS_{\bv}} \leq 1 \\ \bv / \|\bv\|_{\bS_{\bv}} & \text{ otherwise} \end{cases}
    \end{align*}
  \end{enumerate}
  \item Return $\hat{\bu}$ and $\hat{\bv}$, optionally scaled to have (Euclidean) norm 1 
\end{enumerate}
\end{algorithm}
\vspace{-0.1in}\noindent In the final step, $\hat{\bu}$ and $\hat{\bv}$ may be rescaled to have unit norm, as with standard PCA and other regularized variants, but if so, they may no longer be feasible for Problem \eqref{eqn:sfpca}. Despite the non-convexity of the SFPCA problem \eqref{eqn:sfpca}, Algorithm \ref{alg:sfpca} comes with the following strong convergence guarantees: 
\begin{restatable}{theorem}{convergence}
\label{thm:sfpca_convergence}
Under Assumption \ref{ass:general}, Algorithm~\ref{alg:sfpca} has the following properties:
\begin{enumerate}[(i)]
  \item Step 2(a) converges to a stationary point of  
  \begin{equation} \argmin_{\bu \in \overline{\B}^n_{\bS_{\bu}}} \frac{1}{2}\|\bX\bv - \bu\|_2^2 + \lambda_{\bu} P_{\bu}(\bu) + \frac{\alpha_{\bu}}{2}\bu^T\bOmega_{\bu}\bu \label{eqn:sfpca_u_reg}. \end{equation}
  Furthermore, if $P_{\bu}$ is convex, the convergence is monotone, at an $\mathcal{O}(1/K)$ rate, and to a global solution. Step 2(b) converges analogously for $\bv$ and $P_{\bv}$. 
  \item If $P_{\bu}$ is convex, Step 2(a) yields a global solution to \eqref{eqn:sfpca}, considering $\hat{\bv}$ fixed; if $P_{\bu}$ is non-convex, Step 2(a) yields a stationary point for $P_{\bu}$, considering $\hat{\bv}$ fixed. An analogous result holds for $\hat{\bv}$ returned by Step 2(b), with $\hat{\bu}$ considered fixed.
  \item If $P_{\bu}, P_{\bv}$ are both convex, then $(\hat{\bu}, \hat{\bv})$ returned by the SFPCA Algorithm \eqref{alg:sfpca} is both a coordinate-wise global maximum (Nash point) and a stationary point of Problem \eqref{eqn:sfpca}.
\end{enumerate}
\end{restatable}
\noindent We note that the convergence rates associated with steps 2(a) and 2(b) can be further improved to $\mathcal{O}(1/K^2)$ if an accelerated proximal gradient scheme is instead used to solve the $\bu$- or $\bv$-subproblems \citep{Beck:2009}, though monotonicity may be lost. Additionally, in the case where $\alpha_{\bu} = 0$, then subproblem \eqref{eqn:sfpca_u_reg} is solved by normalizing  $\prox_{\lambda_{\bu}P_{\bu}(\cdot)}(\bX\bv)$ and hence converges in a single step. 

Since the SFPCA problem \eqref{eqn:sfpca} is non-convex, the estimates returned by Algorithm \ref{alg:sfpca} depend on the initial values chosen for $\bu$ and $\bv$. In practice, we have found the unregularized singular vectors to provide a robust and easily computed initialization. More complex constraints can be added to SFPCA by incorporating them in the proximal operators applied in steps 2(a) and 2(b) of Algorithm \ref{alg:sfpca}. In particular, we can impose non-negativity constraints of the form considered by \citet{Allen:2011} by incorporating the indicator function of the positive orthant into the penalty functions $P_{\bu}, P_{\bv}$; for many popular penalties, this yields a positive proximal operator with a closed form, \emph{e.g.}, the positive-part operator when the underlying penalty is the lasso. 

Algorithm \ref{alg:sfpca} returns estimates of the leading left and right regularized singular vectors of $\bX$ only. Additional regularized singular vectors can be obtained by iteratively applying Algorithm \ref{alg:sfpca} to a suitably deflated data matrix. In our simulation and case studies in the next two sections, we use Hotelling's subtraction deflation ($\bX := \bX - d \hat{\bu}\hat{\bv}^T$ where $d = \hat{\bu}^T\bX\hat{\bv}$), though the alternative deflation schemes proposed by \citet{Mackey:2008} could be also be used. 

Because Algorithm \ref{alg:sfpca} essentially only requires solving penalized regression problems, it avoids the expensive matrix inversion or eigendecomposition steps common to other regularized PCA variants. For problems with closed-form proximal operators that can be evaluated in linear time, the computational cost of Algorithm \ref{alg:sfpca} is $\mathcal{O}(n^2 + p^2)$, dominated by the cost of multiplication by $\bS_{\bu}$ and $\bS_{\bv}$. As smoothing matrices typically have a banded structure, additional problem-specific improvements are often possible. We also note that randomized methods \citep{Halko:2011} can be used to efficiently obtain estimates of the leading singular vectors of $\bX$ used to initialize $\hat{\bu}, \hat{\bv}$ in Algorithm \ref{alg:sfpca}, thereby avoiding an expensive computation in very large problems.

\begin{table*}
\vspace{-0.13in}
\centering
\scalebox{.85}{
\begin{tabular}{l|lr|r|r|r|r|r|r}
\toprule
\multicolumn{2}{c}{} & & \multicolumn{1}{c}{TWFPCA} & \multicolumn{1}{c}{SSVD} & \multicolumn{1}{c}{PMD} & \multicolumn{1}{c}{SGPCA ($\sigma=1$)} & \multicolumn{1}{c}{SGPCA ($\sigma=5$)} & \multicolumn{1}{c}{SFPCA} \\
\midrule
\multirow{10}{*}{$n=100$} & \multirow{3}{*}{$\bv_{1}$} & TP & - & 0.897 (.004) & 0.568 (.005) & 0.768
  (.008) & 0.820 (.004) & {\bf 0.935} (.004) \\
& & FP & - & 0.323 (.080) & 0.001 (.000) & {\bf 0.006} (.002) & 0.012	(.002)
  & 0.052 (.032) \\
& & r$\angle$ & {\bf 0.153} (.055) & 0.625 (.112) & 2.220 (.035) & 0.726 (.024) &
  0.369 (.007) & 0.189 (.062) \\
\cline{2-9}
& \multirow{3}{*}{$\bv_{2}$}  & TP& - & {\bf 0.783} (.007) & 0.657 (.006) & 0.445 (.010) & 0.005
  (.002) & 0.713 (.008) \\ 
& & FP & - & 0.320 (.080) & 0.106 (.004) & {\bf 0.002} (.001) & 0.257	(.003)
  & 0.047 (.031) \\
& & r$\angle$ & 5.980 (.346) & 0.549 (.105) & 0.597 (.012) & 0.829 (.024) &
  6.150	(.104) & {\bf 0.438} (.094) \\
\cline{2-9}
& \multirow{3}{*}{$\bv_{3}$}  & TP & - & 0.771 (.007) & 0.514 (.007) & 0.499 (.015) &
  0.064	(.014) & {\bf 0.883} (.008) \\
& & FP & - & 0.316 (.079) & 0.066 (.004) & {\bf 0.004} (.002) & 0.128	(.014)
  & 0.054 (.033) \\
& & r$\angle$ & 3.660 (.270) & 0.855 (.131) & 1.270 (.023) & 1.010 (.038) &
  4.000 (.093) & {\bf 0.468} (.097) \\
\cline{2-9}
& & rSE & 0.668 (.003) & 0.760 (.002) & 1.000 (.008) & 0.737 (.009)
  & 0.936 (.017) & {\bf 0.450} (.003) \\
\midrule
\multirow{10}{*}{$n=300$} & \multirow{3}{*}{$\bv_{1}$}  & TP & - & 0.973 (.002) & 0.509 (.003) & 0.921
(.003) & 0.904 (.002) & {\bf 0.987} (.001) \\ 
& & FP & - & 0.322 (.080) & {\bf 0.000} (.000) & 0.005 (.002) & 0.015	(.002)
& 0.068 (.037) \\
& & r$\angle$ & 0.768 (.124) & 0.487	(.099) & 15.700 (.292)& 0.553 (.017) &
0.443 (.011) & {\bf 0.152} (.055) \\
\cline{2-9}
& \multirow{3}{*}{$\bv_{2}$}  & TP & - & 0.919 (.004) & 0.773 (.003) & 0.839 (.004) &
0.011	(.003) & {\bf 0.967} (.003) \\
& & FP & - & 0.319 (.080) & {\bf 0.000} (.000) & 0.038 (.003) & 0.323	(.002)
& 0.048 (.031) \\
& & r$\angle$ & 52.300 (1.02) & 0.428 (.093) & 1.310	(.023) & 0.488 (.024)
& 52.800 (.935) & {\bf 0.320} (.080) \\
\cline{2-9}
& \multirow{3}{*}{$\bv_{3}$}  & TP & - & 0.943 (.003) & 0.530 (.004) & 0.849 (.006) &
0.005	(.002) & {\bf 0.972} (.002) \\
& & FP & - & 0.314 (.079) & {\bf 0.000} (.000) & 0.015 (.003) & 0.212	(.002)
& 0.060 (.035) \\
& & r$\angle$ & 33.100 (.813) & 0.545 (.104) & 5.940 (.089) & 0.631 (.026)
& 34.200 (.543) & {\bf 0.131}	(.051) \\
\cline{2-9}
& & rSE & 1.170 (.002) & 0.790 (.001) & 3.380	(.016) & 0.809 (.005)
& 1.360 (.007) & {\bf 0.655} (.001) \\
\bottomrule
\end{tabular}}
\caption{Performance of various regularized PCA methods for the simulation study described in Section \ref{sec:simulation}. Results are averaged over 50 replicates, with standard errors given in parentheses. For each method, the true positive rate (TP), false positive rate (FP), relative angle compared to that of the SVD (r$\angle$), and relative squared error compared to that of the SVD (rSE) are reported. (TP and FP are not reported for the non-sparse TWFPCA.) The best performing method on each metric is bold-faced. SFPCA consistently outperforms other methods.} 
\label{tab:simulationI}
\end{table*}

\subsection{Selection of Regularization Parameters}
While Algorithm \ref{alg:sfpca} provides an efficient and scalable approach to fitting SFPCA on large data sets, we have not yet addressed the question of tuning various regularization parameters. The presence of four independently chosen tuning parameters -- $\lambda_{\bu}, \lambda_{\bv}, \alpha_{\bu}, \alpha_{\bv}$ -- would appear to be a major drawback of our formulation. Indeed, cross-validation over a four dimensional grid of regularization parameters would pose a significant computational burden. Instead we adapt the strategy of \citet{Huang:2009}, who exploit the connection between two-way FPCA and penalized regression methods to develop an efficient tuning scheme.

In particular, we propose a greedy ``coordinate-wise'' Bayesian Information Criterion (BIC) optimization scheme. We begin by holding the tuning parameters associated with $\bv$ fixed ($\alpha_{\bv}, \lambda_{\bv}$) and choosing $\alpha_{\bu}$ and $\lambda_{\bu}$ to optimize the BIC of the $\bu$-subproblem \eqref{eqn:sfpca_u_reg}. We then hold $\alpha_{\bu}$ and $\lambda_{\bu}$ and optimize the BIC of the $\bv$-subproblem. If these searches are embedded within a  warm-starting scheme for steps 2(a) and 2(b) of Algorithm \ref{alg:sfpca}, this can be achieved with minimal additional computational cost. The degrees of freedom and associated BIC of the $\bu$- and $\bv$-subproblems can be established using the techniques proposed by \citet{Kato:2009} and \citet{Tibshirani:2012}, though we provide an explicit expression for the common case of an $\ell_1$ sparsity penalty:

\begin{restatable}{theorem}{bic} \label{thm:bic}
Suppose $P_{\bu}(\bu) = \|\bu\|_1$. Then an unbiased estimate degrees of freedom of the $\bu$-subproblem \eqref{eqn:sfpca_u_reg} is given by 
\[\widehat{\mathrm{df}}(\hat{\bu}) =  \mathrm{Tr}\left[\left(\bI_{|\mathcal{A}|} +
\alpha_{\bu} \bOmega_{\bu}^{\mathcal{A}}\right)^{-1} \right]\approx \mathrm{Tr}\left[ \bI_{|\mathcal{A}|} -
\alpha_{\bu} \bOmega_{\bu}^{\mathcal{A}} \right]\]
where $\mathcal{A}$ denotes the indices of the estimated non-zero elements of
$\hat{\bu}$ and $\bOmega_{\bu}^{\mathcal{A}}$ denotes the corresponding submatrix of $\bOmega_{\bu}$. Hence, the approximate BIC to be optimized for subproblem \eqref{eqn:sfpca_u_reg} is given by 
\[\mathrm{BIC}(\hat{\bu}) = \log\left[ \frac{1}{n}\|\bX \bv-  \hat{\bu}\|_{2}^{2}\right] + \frac{1}{n} \log(n) \,  \widehat{\mathrm{df}}(\hat{\bu}).\]
\end{restatable}
\noindent One potential shortcoming of our proposed approach is that the greedy search is not guaranteed to converge and may enter an infinite loop as it attempts to optimize the regularization parameters. To address this, non-convergence guards (\emph{e.g.}, a maximum number of steps) may be added, but in our experience, however, the greedy search tends to stabilize quickly and guards against non-convergence are not needed for most problems. As shown in the next two section, this scheme performs well in practice, selecting flexible combinations of sparsity and smoothness in a tractable data-driven manner. 

\section{Simulation Study} \label{sec:simulation}
In this section, we compare the performance of our SFPCA method \eqref{eqn:sfpca} with several competitors including the two-way FPCA (TWFPCA)  method of \citet{Huang:2009}, the sparse SVD method (SSVD) of \citet{Lee:2010}, the penalized matrix decomposition (PMD) of \citet{Witten:2009}, and the sparse generalized PCA (SGPCA) of \citet{Allen:2014}. We simulate data according to the low-rank model $\bX = \sum_{k=1}^K d_k \bu_k \bv_k^T + \bE$ where $E_{ij} \buildrel\textsc{iid}\over\sim \mathcal{N}(0, 1)$. We fix $K = 3$ and $p = 200$ and sample the left singular vectors uniformly $\bv$ from the space of orthogonal matrices. The signal in the right singular vectors $\bv$, each of which have a combination of sparsity and smoothness, takes the form of a sinusoidal pulse. The scale-factors $d_i$, which control the signal-to-noise ratio, vary with the sample size as $d_1 = n / 4, d_2 = n / 5, d_3 = n / 6$.

The SGPCA generalizing operators were constructed using the method suggested by \citet{Allen:2011} with kernel $e^{-d_{ij}^2/\sigma}$ for Chebychev distances between time points $i, j$. The smoothing matrices $\Omega_{\bu}, \Omega_{\bv}$ were fixed as squared second difference matrices. The sparse methods were implemented using an unweighted $\ell_1$-penalty. Tuning parameters for each method were selected using the authors' recommended approach. For SFPCA, the greedy BIC method described above was used. 

Our qualitiative results are shown in Figure \ref{fig:simulationI}, where we see that SFPCA clearly outperforms the competing methods. The non-sparse standard SVD and TWFPCA are not able to successfully localize the sinusoidal pulses in time, while the non-smooth PMD and SGPCA are not able to recover the smooth sinusoidal structure. 

\begin{figure*}[h]
\centering
\includegraphics[width=2.75in]{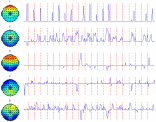}\hspace{4mm}\includegraphics[width=2.75in]{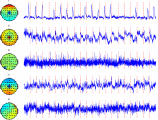}
\caption{EEG Case Study: first five spatial and temporal SFPCA components (left) and ICA components (right). While SFPCA and ICA identify similar structures in the first two components, the temporal sparsity of the SFPCA components makes them more readily interpretable. Additionally, the SFPCA finds structure in the subsequent components that ICA does not identify.}
\label{fig:eeg}
\end{figure*}

Quantitative results are presented in Table \ref{tab:simulationI}, where we report the true positive rate (TP) and false positive rate (FP) for recovering the support of $\bv$, as well as two measures of smoothness, the relative angle and the relative squared error. The relative angle is given by $\text{r}\angle = (1 - |\hat{\bv}^T\bv^*|) / (1 - |\hat{\bv}^T_{\text{SVD}}\bv^*|)$ where $\bv^*$ is the true signal and $\hat{\bv}_{\text{SVD}}$ is the SVD-estimated singular vector; smaller values of $\text{r}\angle$ indicate better performance, with values less than one signifying more accurate estimation than the standard SVD. The relative squared error measures the reconstruction accuracy and is given by $\text{rSE} = \|\bX^* - \hat{\bX}\|_F^2 / \|\bX^* - \hat{\bX}^{\text{3-SVD}}\|_F^2$; smaller values of $\text{rSE}$ indicate better performance, with values less than one signifying more accurate estimation than the standard SVD. (Note that that for both measures, we consider reconstruction of the true mean matrix and true right singular vectors, else it would be impossible to outperform the SVD.) SFPCA consistently outperforms the other regularized PCA methods and, as measured by $\text{r}\angle$ and $\text{rSE}$, the standard SVD. Clearly, SFPCA is able to accurately and adaptively recover principal components with complex structure, yielding improved statistical performance. As we will see in the next section, the structured principal components yielded by SFPCA are also more interpretable, making SFPCA a useful tool for exploratory data analysis and scientific model construction.

\section{Case Study: EEG Data} \label{sec:eeg}

We close with an application of SFPCA to a sample of  electoencephalography (EEG) data taken from the UCI Machine Learning Repository \citep{Dua:2017}.\footnote{\url{https://archive.ics.uci.edu/ml/datasets/eeg+database}} These data consist of $n = 57$ EEG channels with corresponding scalp locations and $p = 5376$ time points, corresponding to 21 epochs of 256 time points each. Back-block pattern recognition techniques, especially independent components analysis (ICA), are commonly applied to EEG data to separate sources from the limited channel recordings, find major spatial patterns and corresponding temporal activity patterns, find artifacts in the data, and develop visualizations \citep{Makeig:1995}. SFPCA was applied to the EEG recording from the first alcoholic subject over epochs relating to non-matching stimuli. The spatial smoothing matrix, $\bOmega_{\bu}$, was specified as the weighted squared second differences matrix using spherical distances between the recording channel locations and the temporal smoothing matrix, $\bOmega_{\bv}$, was taken as the matrix of squared second differences. Tuning parameters for SFPCA were selected using the greedy scheme described above.

In Figure \ref{fig:eeg}, we compare the SFPCA results with those obtained from the FastICA method \citep{Hyvarinen:2000}. At a high level, the patterns identified by SFPCA and ICA are similar, identifying the same major temporal patterns and spatial source localization, but the SFPCA results are much more directly interpretable. The improvements afforded by SFPCA are clearly seen by comparing the first two components, where the spatial patterns are similar but SFPCA identifies a much more structured temporal pattern. Furthermore, SFPCA is able to identify more signals: the third SFPCA vectors identify a singular ``pulse'' which is spatially and temporally localized, while the third ICA component has no discernable structure. 

Interestingly, the greedy BIC scheme consistently selects $\lambda_{\bu} = 0$, suggesting that no sparsity in the EEG channels is needed. Conversely, the greedy scheme consistently selected non-zero smoothing and temporal sparsity parameters for each of the first five SFPCA components ($\alpha_{\bu} \in [10, 12]$, $\alpha_{\bv} \in [0.5, 10]$, $\lambda_{\bv} \in [1, 2.5]$), indicating that our method is able to flexibly choose the optimal degree of smoothness and sparsity for recovering major patterns in the data.

\section{Discussion}

We have proposed SFPCA, a flexible yet coherent approach to sparsity- and smoothness-regularized PCA. This flexibility gives SFPCA the ability to adapt to the types and amounts of regularization appropriate for a given problem in a data-driven manner. SFPCA unifies much of the existing literature on regularized PCA and allows for as-of-yet-unexplored generalizations by varying the penalty functions and smoothing matrices. In our simulation and case studies, SFPCA exhibits superior statistical performance and improved interpretability. As special cases of SFPCA have been shown to lead to consistent estimation of principal components, even in the high-dimensional context \citep{Silverman:1996,Shen:2013}, we conjecture that the general SFPCA framework also yields consistent estimates, an interesting topic for future research. 

The advantages of SFPCA are not purely theoretical, however: Algorithm \ref{alg:sfpca} provides a framework for solving the SFPCA Problem, which is fast and scalable for general problems, while also easily modified to take advantage of additional computational efficiencies afforded by specific problems. As shown in Theorem \ref{thm:sfpca_convergence}, Algorithm \ref{alg:sfpca} enjoys attractive convergence properties despite its inherent non-convexity. Additionally, the greedy BIC scheme we have proposed allows for computationally efficient determination of regularization parameters.  \textsc{Matlab} scripts implementing SFPCA are available from the first author's website. Supplemental materials for this paper including proofs and additional experiments are available at \url{https://arxiv.org/abs/1309.2895}.

The advantages of SFPCA demonstrated here suggest additional lines of research, including extensions to the multi-way (tensor) context using the framework established by \citet{Allen:2012} or to other widely-used multivariate analysis techniques, such as partial least squares (PLS), canonical correlation analysis (CCA), and linear discriminant analysis (LDA).

\section{References}
\printbibliography[heading=none]
\end{refsection}
\begin{refsection}
\onecolumn
\appendix

\section*{Supplementary Materials}
\section{Proofs}

Before proving the major results stated in the main body of the paper, we give three lemmas: 
\begin{lemma} \label{lem:subgrad_homo}
Suppose $f(\bx): \R^p \to \R_{\geq 0}$ is a non-negative function and is positive homogeneous of order one, \emph{i.e.}, $f(c\bx) = cf(\bx)$ for all $\bx \in \R^p$ and all $c \geq 0$. Then, if $\tilde{\nabla}f(\bx)$ is a sub-gradient of $f$ at $\bx$, then $\tilde{\nabla}f(c\bx)$ is also a sub-gradient for all $c \geq 0$. 
\end{lemma}
\begin{proof} This follows immediately from the definition of a sub-gradient and the assumption of positive homogeneity. If $\tilde{\nabla}f(\bx)$ is a sub-gradient of $f$ and $\bx$, then we have
\[f(\by) \geq f(\bx) + \tilde{\nabla}f(\bx)^T(\by - \bx) \quad \forall y \in \R^p\]
Substitute $\bx \to c\bx$ and $\by \to c\by$ for arbitrary $c \geq 0$ to obtain \[f(c\by) \geq f(c\bx) + \tilde{\nabla}f(c\bx)^T(c\by - c\bx) \quad \forall y \in \R^p.\] Direct simplification yields
\begin{align*}
    f(c\by) &\geq f(c\bx) + \tilde{\nabla}f(c\bx)^T(c\by - c\bx) \quad \forall \by \in \R^p\\
    c f(\by) &\geq cf(\bx) + c\tilde{\nabla}f(c\bx)^T(\by - \bx) \quad \forall \by \in \R^p\\
    f(\by) &\geq f(\bx) + \tilde{\nabla}f(c\bx)^T(\by - \bx) \quad \forall \by \in \R^p
\end{align*}
which implies that $\tilde{\nabla}f(c\bx)$ is also a sub-gradient of $f$ and $\bx$.
\end{proof}

\begin{lemma} \label{lem:sfpca_kkt}
Suppose $(\bu, \bv)$ are a global maximum of the SFPCA Problem \eqref{eqn:sfpca}. Then $(\bu, \bv)$ satisfy the following Karush-Kuhn-Tucker (KKT) conditions: 
\begin{align*}
  \bX\bv - \lambda_{\bu}\tilde{\nabla}_{\bu}P_{\bu}(\bu) - 2\gamma_{\bu}\bS_{\bu}\bu &= 0 \quad (\bu\text{-stationarity})\\
  \gamma_{\bu}(\|\bu\|_{\bS_{\bu}} - 1) &= 0 \quad (\bu\text{-complementary slackness})\\
  \bu^T\bX - \lambda_{\bv}\tilde{\nabla}_{\bv}P_{\bv}(\bv) - 2\gamma_{\bv}\bS_{\bv}\bv &= 0 \quad (\bv\text{-stationarity})\\
  \gamma_{\bv}(\|\bv\|_{\bS_{\bv}} - 1) &= 0 \quad (\bv\text{-complementary slackness})\\
  \gamma_{\bu} &\geq 0 \quad (\bu\text{-dual feasibility}) \\
  \gamma_{\bv} &\geq 0 \quad (\bv\text{-dual feasibility}) \\
  \|\bu\|_{\bS_{\bu}} &\leq 1 \quad (\bu\text{-primal feasibility}) \\
  \|\bv\|_{\bS_{\bu}} &\leq 1 \quad (\bv\text{-primal feasibility})
\end{align*}
where $\gamma_{\bu}$ and $\gamma_{\bv}$ are the dual variables associated with the inequality constraints of the SFPCA Problem \eqref{eqn:sfpca} and $\tilde{\nabla}f(\bx)$ denotes an arbitrary sub-gradient of $f$ and $\bx$: that is, any value satisfying $f(\by) \geq f(\bx) + \tilde{\nabla}f(\bx)^T(\by - \bx)$ for all $\by \in \R$. 
\end{lemma}
\begin{proof} Despite the non-convexity of the SFPCA Problem \eqref{eqn:sfpca}, many of the classical results of convex analysis, including the KKT conditions, can be established for local minima under additional assumptions. Chapter 5 of \citet{Bertsekas:2003} gives an elegant presentation of these results. In particular, we note that the SFPCA Problem \eqref{eqn:sfpca} satisfies their CQ5c, a variant of Slater's condition \citep{Slater:1950}, for any local maximum as the point $(\bzero, \bzero)$ is clearly strictly feasible. Additionally, we note that the feasible set $\overline{\B}^n_{\bS_{\bu}} \times \overline{\B}^p_{\bS_{\bv}}$ is clearly \emph{regular} in their sense of having well-behaved normal and (polar) tangent cones (see, \emph{e.g.}, their Definition 4.6.3). Since any global optimum must be a local optimum, the desired result follows. (The top right portion of their Figure 5.5.2 of \citet{Bertsekas:2003} is useful in following their presentation.)
\end{proof}

\begin{lemma} \label{lem:sfpca_u_prob_kkt}
Suppose $P_{\bu}: \R^p \to \R_{\geq 0}$ is a positive-homogeneous function of order one. Let 
\begin{equation} \label{eqn:kkt_lemma_eqn2}
\bu^* = \begin{cases} \hat{\bu} / \|\hat{\bu}\|_{\bS_{\bu}} & \text{ where } \|\bu\|_{\bS_{\bu}} > 0 \\ \bzero & \text{ otherwise } \end{cases} \quad \text{ where $\hat{\bu}$ is a stationary point of } \quad \argmin_{\bu \in \R^n} \frac{1}{2}\|\bX\bv - \bu\|_2^2 + \lambda_{\bu}P_{\bu}(\bu) + \frac{\alpha}{2}\bu^T\bOmega_{\bu}\bu 
\end{equation} Then $\bu^*$ is a stationary point of 
\begin{equation}
  \argmax_{\bu \in \overline{\B}^n_{\bS_{\bu}}} \bu^T\bX\bv - \lambda_{\bu}P_{\bu}(\bu) \label{eqn:kkt_lemma_eqn1}
\end{equation} (Note that Problem \eqref{eqn:kkt_lemma_eqn2} is Problem \eqref{eqn:sfpca_u_reg} from the main text restated here for convenience.) Additionally, if $P_{\bu}$ is convex, then $\hat{\bu}$ and $\bu^*$ are global optima of their respective subproblems. 
\end{lemma}
\begin{proof} We note that this proof follows the proof of Theorem 2 of \citet{Allen:2014}. Following the proof of Lemma \ref{lem:sfpca_kkt} holding $\bv$ fixed (and feasible), we have the following KKT conditions for Problem \eqref{eqn:kkt_lemma_eqn1}:
\begin{align*}
  2\gamma_{\bu}\bS_{\bu}\bu^* - \bX\bv + \lambda_{\bu}\tilde{\nabla}P_{\bu}(\bu^*) &= 0 \quad \text{(stationarity)} \\
  \gamma_{\bu}(\|\bu^*\|_{\bS_{\bu}} - 1) &= 0 \quad \text{(complementary slackness)}.
\end{align*}
Similarly, the KKT conditions for $\hat{\bu}$ in Problem \eqref{eqn:kkt_lemma_eqn2} yield: 
\[0 = -(\bX\bv - \hat{\bu}) + \lambda_{\bu}\tilde{\nabla}P_{\bu}(\hat{\bu}) + \alpha\bOmega_{\bu}\hat{\bu} \implies \bS_{\bu}\hat{\bu} - \bX\bv + \lambda_{\bu}\tilde{\nabla}P_{\bu}(\hat{\bu}) = 0\]
where $\hat{\bu} + \alpha\Omega\hat{\bu} = \bS_{\bu}\hat{\bu}$. Comparing the stationarity conditions for $\bu^*$ and $\hat{\bu}$, we see that they are equivalent up to the $2\gamma_{\bu}$ term. 

Let $\tilde{\bu} = \hat{\bu} / 2\gamma_{\bu}$. Then the KKT conditions of Problem \eqref{eqn:kkt_lemma_eqn2} imply: 
\begin{align*}
 0 &= \bS_{\bu}\hat{\bu} - \bX\bv + \lambda_{\bu}\tilde{\nabla}P_{\bu}(\hat{\bu}) \\ 
 &= 2\gamma_{\bu}\bS_{\bu}\tilde{\bu} - \bX\bv + \lambda_{\bu}\tilde{\nabla}P_{\bu}(2\gamma_{\bu}\tilde{\bu}) \\
 &= 2\gamma_{\bu}\bS_{\bu}\tilde{\bu} - \bX\bv + \lambda_{\bu}\tilde{\nabla}P_{\bu}(\tilde{\bu})
\end{align*}
where the constant of $2\gamma_{\bu}$ appearing in the sub-gradient could be removed using Lemma \ref{lem:subgrad_homo}. From this we see that $\tilde{\bu}$ satisfies the stationarity conditions for Problem \eqref{eqn:kkt_lemma_eqn1}. Hence, if we take $(\bu^*, \gamma_{\bu}) = (\hat{\bu}/\|\hat{\bu}\|_{\bS_{\bu}}, \|\hat{\bu}\|_{\bS_{\bu}} / 2)$, we have a solution to the KKT conditions for Problem \eqref{eqn:kkt_lemma_eqn1}, implying that we have a local solution. Additionally, if $P_{\bu}(\cdot)$ is convex, then Problem \eqref{eqn:kkt_lemma_eqn1} is concave, so the KKT conditions imply global optimality.

More intuitively, if we compare the stationarity conditions for $\bu^*$  and $\hat{\bu}$ directly, we see that they differ only by the leading constant factor of $2\gamma_{\bu}$, suggesting that $\bu^* \propto \hat{\bu}$. Since we know $\bu^*$ is a unit-vector under the $\bS_{\bu}$-norm, we can guess $\bu^* = \hat{\bu}/\|\hat{\bu}\|_{\bS_{\bu}}$, which, when substituted into the KKT conditions, yields $\gamma_{\bu} = \|\hat{\bu}\|_{\bS_{\bu}}$.
\end{proof}

\noindent With these results in hand, we are now ready to prove the main results of our paper, which we restate here for convenience.

\kkt*
\begin{proof}[Proof of Theorem \ref{thm:kkt}] Throughout the following, we continue the notation used in the proof of Lemma \ref{lem:sfpca_u_prob_kkt} and let $(\bu^*, \bv^*)$ denote solutions to the SFPCA problem \eqref{eqn:sfpca}, while $\hat{\bu}$ and $\hat{\bv}$ denote solutions to Problem \eqref{eqn:kkt_lemma_eqn2} and its analogue in $\bv$: 
\[\hat{\bv} = \argmin_{\bv \in \R^p} \frac{1}{2}\|\bu^T\bX - \bv\|_2^2 + \lambda_{\bv}P_{\bv}(\bv) + \frac{\alpha}{2}\bv\bOmega_{\bv}\bv\]
\begin{enumerate}[{Part} (i)] 
\item From Lemma \ref{lem:sfpca_u_prob_kkt}, we have that $\bu^* = 0$ if and only if $\hat{\bu} = 0$. The KKT conditions for Problem \eqref{eqn:kkt_lemma_eqn2} show this occurs only when \[\lambda_{\bu}\tilde{\nabla}P_{\bu}(\bzero) = \bX\bv.\] Hence, for any fixed  $\bv$, we can find a value $\lambda_{\bu}^{\bv, \max}$, such that $\lambda_{\bu} \geq \lambda_{\bu}^{\bv, \max}$ yields an all-zero solution. Taking the maximum over all $\bv \in \overline{\B}^p_{\bS_{\bv}}$, we obtain $\lambda_{\bu}^{\max}$ as desired. An analogous result holds for $\lambda_{\bv}^{\max}$. 

Additionally, we note that if $\bu = \bzero$, then $\bv = \bzero$ satisfies the KKT conditions given in Lemma \ref{lem:sfpca_kkt} and hence is a solution. Putting these together, we note that if $\lambda_{\bu} \geq \lambda_{\bu}^{\max}$ or if $\lambda_{\bv} \geq \lambda_{\bv}^{\max}$, then $(\bu^*, \bv^*) = (\bzero, \bzero)$, as desired. 
\item Now, we assume $\lambda_{\bu} < \lambda_{\bu}^{\max}$ and $\lambda_{\bv} < \lambda_{\bv}^{\max}$, so $\bu^* \neq \bzero$ and $\bv^* \neq 0$, and $\alpha_{\bu}, \alpha_{\bv} > 0$. By the $\bu$-stationary term of the KKT conditions given in Lemma \ref{lem:sfpca_kkt}, it is clear that $\bu^*$ depends on both $\lambda_{\bu}$ and $\alpha_{\bu}$, by way of $\bS_{\bu}$, as well as $\bv^*$. A similar argument shows that $\bv^*$ depends on both $\lambda_{\bv}$ and $\alpha_{\bv}$ as well as $\bu^*$, so transitively both $\bu^*$ and $\bv^*$ depend on all (non-zero) regularization parameters. 
\item Consider the $\bu$-complimentary slackness condition given in Lemma \ref{lem:sfpca_kkt}, which implies that $\gamma_{\bu} > 0$ if and only if $\|\bu^*\|_{\bS_{\bu}} = 1$. In the proof of Lemma \ref{lem:sfpca_u_prob_kkt}, we showed that solutions to the SFPCA KKT conditions are of the form $(\bu^*, \gamma_{\bu}) = (\hat{\bu} / \|\hat{\bu}\|_{\bS_{\bu}}, \|\hat{\bu}\|_{\bS_{\bu}}/2)$. Hence, $\gamma_{\bu} = 0$ if and only if $\hat{\bu} = \bzero$, which, by Part (i), occurs when $\lambda_{\bu} \geq \lambda_{\bu}^{\max}$ or $\lambda_{\bv} \geq \lambda_{\bv}^{\max}$. Putting this together, if the $(\bu^*, \bv^*)$ are non-zero, then the boundary conditions must hold with $\|\bu^*\|_{\bS_{\bu}} = \|\bv^*\|_{\bS_{\bv}} = 1$. 
\item As shown in Part (iii), for non-trivial solutions we have $\|\bu^*\|_{\bS_{\bu}} = \|\bv^*\|_{\bS_{\bv}} = 1$, so the SFPCA problem does not suffer from scale non-identifiability: that is, if $(\bu, \bv)$ is a solution, we do not have additional solutions of the form $(c\bu, c^{-1}\bv)$ for $c > 0$. If $P_{\bu}, P_{\bv}$ are even functions (that is $P_{\bu}(-\bu) = P_{\bu}(\bu)$ and $P_{\bv}(-\bv) = P_{\bv}(\bv)$ for all $\bu, \bv$), the SFPCA problem still has a sign non-identifiability.
\end{enumerate} 
\end{proof}

\equivalence*
\begin{proof}[Proof of Theorem \ref{thm:equivalence}] We establish the equivalence of several cases of SFPCA with approaches previously proposed in the literature. 
\begin{enumerate}[{Part} (i)]
\item $\lambda_{\bu} = \lambda_{\bv} = \alpha_{\bu} = \alpha_{\bv} = 0$. In this case, the SFPCA Problem \eqref{eqn:sfpca} simplifies to 
\[\argmax_{\substack{\bu \in \R^n: \bu^T\bu \leq 1 \\ \bv \in \R^p : \bv^T\bv \leq 1}} \bu^T\bX\bv\] which we recognize as the Singular Value Problem \eqref{eqn:svp}, which defines standard PCA.
\item $\lambda_{\bu} = \alpha_{\bu} = \alpha_{\bv} = 0$. The SPCA estimator of \citet{Shen:2008} is given by \[\bv^* = \hat{\bv} / \|\hat{\bv}\| \text{ where } \hat{\bu}, \hat{\bv} = \argmin_{\substack{\bu \in \overline{\B}^n \\ \bv \in \R^p}} \|\bX - \bu\bv^T\|_F^2 + \lambda_{\bv}P_{\bv}(\bv)\] Taking the KKT conditions with respect to $\bv$, we obtain: 
\[\bX^T\hat{\bu} - \hat{\bv} - \lambda_{\bv}\tilde{\nabla}P_{\bv}(\hat{\bv}) = 0\] Comparing this to the KKT conditions for SFPCA derived in Lemma \ref{lem:sfpca_kkt} with $\lambda_{\bu} = \alpha_{\bu} = \alpha_{\bv} = 0$, 
\begin{align*}
  \bX^T\hat{\bu} - 2\gamma_{\bv}\hat{\bv} - \lambda_{\bv}\tilde{\nabla}P_{\bv}(\hat{\bv}) &= 0 \\ 
  \gamma_{\bv}(\|\hat{\bv}\| - 1) &= 0,
\end{align*}
we see that the only difference is the factor of $2\gamma_{\bv}$ in the stationarity conditions. As before, we define $\hat{\bv} = 2\gamma_{\bv}\tilde{\bv}$ and re-write the SPCA stationarity conditions as 
\[0 = \bX^T\hat{\bu} - \hat{\bv} - \lambda_{\bv}\tilde{\nabla}P_{\bv}(\hat{\bv}) = \bX^T\hat{\bu} - 2\gamma_{\bv}\tilde{\bv} - \lambda_{\bv}\tilde{\nabla}P_{\bv}(2\gamma{\bv}\tilde{\bv}) = \bX^T\hat{\bu} - 2\gamma_{\bv}\tilde{\bv} - \lambda_{\bv}\tilde{\nabla}P_{\bv}(\tilde{\bv}), \]
where the final equality follows from Lemma \ref{lem:subgrad_homo}. This clearly matches the $\bv$-stationarity condition for SFPCA and the scaling step implies the complementary slackness condition holds, showing the two solutions are equivalent. 
\item $\alpha_{\bu} = \alpha_{\bv} = 0$. In this case, the SFPCA Problem \eqref{eqn:sfpca} simplifies to \[\argmax_{\substack{\bu \in \R^n: \bu^T\bu \leq 1 \\ \bv \in \R^p : \bv^T\bv \leq 1}} \bu^T\bX\bv - \lambda_{\bu} P_{\bu}(\bu) - \lambda_{\bv}P_{\bv}(\bv)\] which is clearly equivalent to the sparse GPCA method of \citet[Equation 6]{Allen:2014} with the generalizing operators $\bQ, \bR$ both set equal to identity matrices. For non-convex problems such as SFPCA \eqref{eqn:sfpca}, it is not always the case that constraints can be re-written as Lagrange multipliers and penalty functions; conditions under which this is possible are discussed in Chapter 5 of \citet{Bertsekas:2003} and do indeed apply here. (See also the discussion in the proof of Lemma \ref{lem:sfpca_kkt}.)
\item $\lambda_{\bu} = \lambda_{\bv} = \alpha_{\bu} = 0$. \citet{Huang:2008} consider a penalized regression formulation of FPCA: \[\frac{1}{2}\|\bX - \bu\bv^T\|_F^2 + \alpha \bu^T\bu \bv^T\bOmega_{\bv}\bv.\] They show that $\bv$ of this formulation is equivalent to (a discretization of) the earlier FPCA formulation of \citet{Silverman:1996}: 
\[\argmax_{\bv \in \R^p} \bv^T\bX^T\bX\bv \quad \text{ subject to } \quad \bv^T\bS_{\bv}\bv = 1\] We compare this to our SFPCA formulation with only $\alpha_{\bv}$ non-zero: 
\[\argmax_{\bu, \bv} \bu^T\bX\bv \quad \text{ subject to } \quad \bu^T\bu \leq 1 \text{ and } \bv^T\bS_{\bv}\bv \leq 1\]
Examination of the KKT conditions reveals that, for given $\bv$, the above criterion is maximized by taking $\bu = \bX\bv / \sqrt{\bv^T\bX^T\bX\bv}$. Substituting this into the above, we see that SFPCA simplifies to \[\argmax_{\bv} \sqrt{\bv^T\bX^T\bX\bv} \quad \text{ subject to } \quad \bv^T\bS_{\bv}\bv \leq 1\] As shown in Theorem \ref{thm:kkt}, the constraint must hold tightly (since there are no sparsity penalties), and the $\sqrt{\cdot}$ transform is monotonic, so this is clearly equivalent to the FPCA formulation of \citet{Silverman:1996}, which establishes the desired equivalence for the right singular vectors. For the left singular vectors, \citet{Huang:2008} show that the solution to their FPCA formulation is obtained by an iterative method containing the update $\bu := \bX\bv / \|\bv\|_{\bS_{\bv}}$; this is exactly the same as our expression for $\bu$ \emph{modulo} a normalizing factor. 
\item $\lambda_{\bu} = \lambda_{\bv} = 0$. \citet{Huang:2009} consider two-way FPCA as: 
\[\argmax_{\bu \in \R^n, \bv \in \R^p} \bu^T\bX\bv - \frac{\bu^T(\bI + \bOmega_{\bu})\bu \cdot \bv^T(\bI + \bOmega_{\bv})\bv}{2}\] This gives stationarity conditions of the form $\bu \propto (\bI + \bOmega_{\bu})^{-1}\bX\bv$ and $\bv \propto (\bI + \bOmega_{\bv})^{-1}\bX^T\bv$. (Note that \citet{Huang:2009} define their smoothing matrices $\bS_{\bu}$, $\bS_{\bv}$ as the multiplicative inverses of the definitions we use.)
With $\lambda_{\bu} = \lambda_{\bv} = 0$, the SFPCA KKT conditions derived in Lemma \ref{lem:sfpca_kkt} simplify to: 
\begin{align*}
  \bX\bv^* - 2\gamma_{\bu}\bS_{\bu}\bu^* &= 0 \\
  \gamma_{\bu}(\|\bu^*\|_{\bS_{\bu}} - 1) &= 0 \\ 
  \bX^T\bu^* - 2\gamma_{\bv}\bS_{\bv}\bv^* &= 0 \\
  \gamma_{\bv}(\|\bv^*\|_{\bS_{\bv}} - 1) &= 0
\end{align*}
From these, we find $\bu^* \propto \bS_{\bu}^{-1}\bX\bv^*$ and $\bv^* \propto \bS_{\bv}^{-1}\bX^T\bu^*$, which clearly match the two-way FPCA stationary conditions if we take $\alpha_{\bu} = \alpha_{\bv} = 1$, thereby establishing the desired equivalence. We note, however, that the scaling factors used by SFPCA and the method of \citet{Huang:2009} are different, as we take $\bu = \bS_{\bu}^{-1}\bX\bv / \bv^T\bX^T\bS_{\bu}^{-1}\bX\bv$ while they take $\bu = \bS_{\bu}^{-1}\bX\bv / \bv^T\bS_{\bv}\bv$ and similarly for the $\bv$-normalization. This change in scaling is essentially cosmetic, as it does not effect the direction or relative weights of the estimated principal components. \qedhere
\end{enumerate}
\end{proof}

\convergence*
\begin{proof}[Proof of Theorem \ref{thm:sfpca_convergence}] We first note that the $\bu$-subproblem \eqref{eqn:sfpca_u_reg} can be re-written as 
\[\argmin_{\bu \in \overline{\B}^n_{\bS_{\bu}}} \frac{\bu^T\bS_{\bu}\bu}{2} - \bu^T\bX\bv + \lambda_{\bu}P_{\bu}(\bu) = \argmin_{\bu \in \R^n} \underbrace{\frac{\bu^T\bS_{\bu}\bu}{2} - \bu^T\bX\bv}_{\text{smooth}} + \underbrace{\lambda_{\bu}P_{\bu}(\bu) + \iota_{\overline{\B}^n_{\bS_{\bu}}}(\bu)}_{\text{non-differentiable}}\] where $\iota$ represents the (infinite) indicator of the feasible set: that is, $\iota_{\mathcal{X}}(x)$ is zero if $x$ is an element of $\mathcal{X}$ and (positive) infinity otherwise.  We note that the use of the indicator function here is justified despite possible non-convexity because it always holds as a tight constraint since have a feasible point at $\bzero$. 

The first (smooth) term is strictly and strongly convex, since $\bS_{\bu} \succ 0$ by construction, and has a continuous gradient whose Lipschitz constant is given by the leading eigenvalue of $\bS_{\bu}$. We will make repeated use of the proximal mapping of the non-differentiable term, $\lambda_{\bu}P_{\bu} + \iota_{\overline{\B}^n_{\bS_{\bu}}}$, given by $\prox_{f}(\bx) = \argmin_{\bz} \frac{1}{2}\|\bx - \bz\|_2^2 + f(\bz)$ for a given function $f$. For convex functions, the existence and uniqueness of the proximal mapping follow immediately from the properties of strongly convex functions; the properties of the proximal mapping were studied for a wide class of so-called \emph{prox regular} non-convex functions by \citet{Poliquin:1996a,Poliquin:1996b}. Where the proximal mapping is not unique, any minimizer can be used in Algorithm \ref{alg:sfpca}. \citet{Gong:2013} give proximal operators for a range of widely-used convex and non-convex penalty functions. 

We note a general result for any $f$ satisfying the second part of Assumption \ref{ass:general} (positive-homogeneity): 
\[\prox_{f + \iota_{\overline{\mathbb{B}}^n_{\bS_{\bu}}}}(\bx) = \argmin_{\bz} \frac{1}{2}\|\bx - \bz\|_2^2 + f(\bz) + \iota_{\overline{\mathbb{B}}^n_{\bS_{\bu}}}(\bz) = \argmin_{\bz \in \overline{\mathbb{B}}^n_{\bS_{\bu}}} \frac{1}{2}\|\bx - \bz\|_2^2 + f(\bz) = \proj_{\overline{\mathbb{B}}^n_{\bS_{\bu}}}(\prox_f(\bx))\]
where $\proj_{\mathcal{X}}(\bx)$ denotes the projection of $\bx$ onto $\mathcal{X}$. This follows from Theorem 4 of \citet{Yu:2013} where we take $h(\cdot) = \iota_{\R_{> 1}}$ which is clearly an increasing function, and $\|\bx\| = \bx^T\bS_{\bu}\bx$. (See also  Corollary 1 of \citet{Yu:2013}.) Since we assume positive homogeneity of $f$, it is in the class of functions covered by that theorem and the desired result holds. 
\begin{enumerate}[{Part} (i)]
\item Step 2(a) of Algorithm \ref{alg:sfpca} is a standard proximal gradient iteration with fixed step-size applied to the $\bu$-subproblem \eqref{eqn:sfpca_u_reg}. If $P_{\bu}$ is convex, then monotone $\mathcal{O}(1/K)$ convergence to a global solution follows from well-known results on proximal gradient methods: see, \emph{e.g.}, Theorems 10.21 ($\mathcal{O}(1/K)$ convergence), 10.23 (Fej\'er Monotonicity), and 10.24 (convergence to a global optimum) of \citet{Beck:2017}. If $P_{\bu}$ is non-convex, convergence to a stationary point follows from Theorem 10.15(d) of \citet{Beck:2017}. Additionally, we note that, even in the nonconvex setting, step 2(a) monotonically decreases the objective function of the $\bu$-subproblem \eqref{eqn:sfpca_u_reg} \citet[Theorem 10.15(a)]{Beck:2017}.
\item This follows immediately from Lemma \ref{lem:sfpca_u_prob_kkt} and Part (i). 
\item We note that Algorithm \ref{alg:sfpca} can be considered a block-coordinate ascent algorithm for the SFPCA problem \eqref{eqn:sfpca}, where a proximal gradient scheme is used to solve each subproblem. While SFPCA \eqref{eqn:sfpca} is non-concave, it is block bi-concave in $\bu$ and $\bv$, allowing for certain convergence results to be used. In particular, for $P_{\bu}, P_{\bv}$ both convex, we can use the results of \citet[Theorem 4.7]{Gorski:2007} to establish convergence to a so-called Nash point (coordinate-wise optimum) satisfying
\begin{align*}
    f(\bu, \bv^*) \leq f(\bu^*, \bv^*) &\quad \text{ for all } \bu \in \overline{\B}^n_{\bS_{\bu}} \\
    f(\bu^*, \bv) \leq f(\bu^*, \bv^*) &\quad \text{ for all } \bv \in \overline{\B}^p_{\bS_{\bv}}
\end{align*}
where $f(\bu, \bv)$ is the SFPCA objective $f(\bu, \bv) = \bu^T\bX\bv - \lambda_{\bu}P_{\bu}(\bu) - \lambda_{\bv}P_{\bv}(\bv)$. (Theorem 2.3 of \citet{Xu:2013} generalizes this approach to approximate solutions of the subproblem, at the cost of requiring strong convexity.)

\quad To show that the output of Algorithm \ref{alg:sfpca} is also a stationary point, we use the regularity analysis of \citet{Tseng:2001}: in particular, we note that the smooth part of our objective ($f_0(\bu, \bv) = \bu^T\bX\by$) is (G\^{a}teaux-)differentiable everywhere, so Tseng's assumption A1 holds.\footnote{Tseng's treatment of constraints is somewhat unclear here, but we incorporate the unit ellipse constraints as indicator functions in the non-convex penalty portion of the problem, as discussed above, so $\text{dom}\, f_0 = \R^{n} \times \R^p$.} This establishes regularity everywhere, including at each coordinate-wise maximum, which implies each Nash point is also a stationary point. \qedhere
\end{enumerate}
\end{proof}

\noindent We conjecture, but do not prove here, a generalization of the above: if $P_{\bu}$ or $P_{\bv}$ are non-convex, Algorithm \ref{alg:sfpca} is still guaranteed to converge to a coordinate-wise local maximum (local Nash point). \citet{Xu:2017} prove a related result, establishing convergence to a critical point, but where only a single gradient step is taken instead of fully solving the $\bu$- and $\bv$-subproblems as we do in Algorithm \ref{alg:sfpca}. 

Additionally, we note that experimental evidence suggests neither positive-homogeneity nor convexity of $P_{\bu}, P_{\bv}$ are required for convergence of Algorithm \ref{alg:sfpca}, though we are unable to provide a full proof. Similar results have been previously demonstrated for coordinate descent schemes applied to related problems \citep[Theorem 4]{Mazumder:2011} \citep[Proposition 1]{Breheny:2011} \citep[Theorem 3.1]{Xu:2017}, though they do not consider constraints and require a quadratic smooth term which we do not have here. 

Finally, we note that in Step 2(a) of Algorithm \ref{alg:sfpca}, we self-normalize $\bu$ under the $\bS_{\bu}$-norm in order to obtain $\hat{\bu}$ at each step. Algorithmically, this can be considered a projected gradient scheme, where projection is required to ensure feasibility at each step. In the non-sparse case ($\lambda_{\bu} = 0$), this update has the closed form $\hat{\bu} = \bS_{\bu}^{-1}\bX\bv / \|\bS_{\bu}^{-1}\bX\bv\|_{\bS_{\bu}} = \bS_{\bu}^{-1}\bX\bv / \|\bX\bv\|_{\bS_{\bu}^{-1}}$, which is closely related to the updates in the two-way functional PCA algorithm of \citet{Huang:2009}, but using a different normalization. As \citet{Huang:2009} discuss, this is equivalent to the more standard ``half-smoothing'' approach popularized by \citet{Silverman:1996}. (As \citet{Allen:2013} discusses, this equivalence does not extend straightforwardly to the higher-order array (tensor) context.)

\bic*
\begin{proof}[Proof of Theorem \ref{thm:bic}] Consider the $\bu$-update with $\ell_1$-penalization \citep{Tibshirani:1996}. In this case, the $\bu$-subproblem \eqref{eqn:sfpca_u_reg} is essentially a \emph{generalized} elastic net problem, \citep{Zou:2005} which can be analyzed using the techniques of \citet{Tibshirani:2012}. In particular, we re-write Problem \eqref{eqn:sfpca_u_reg} as a lasso problem with an augmented design matrix: 
\[\argmin_{\bu} \frac{1}{2}\left\|\begin{pmatrix} \bX\bv \\ \bzero \end{pmatrix} - \begin{pmatrix} \bI \\ \chol(\alpha_{\bu}\bOmega_{\bu}) \end{pmatrix}\bu \right\|_2^2 + \lambda_{\bu}\|\bu\|_1\] where $\tilde{\bX}$ is the augmented design matrix $\tilde{\bX} = \begin{pmatrix} \bI & \chol(\alpha_{\bu}\Omega_{\bu})^T \end{pmatrix}^T$. Then the degrees of freedom are given by \[\text{df} = \mathbb{E}\left[\Tr\left((\tilde{\bX}_{\mathcal{A}}^T\tilde{\bX}_{\mathcal{A}})^{-1}\right)\right].\] Note that the general form of their estimator is $\Tr(\bX_{\mathcal{A}}(\tilde{\bX}_{\mathcal{A}}^T\tilde{\bX}_{\mathcal{A}})^{-1}\bX_{\mathcal{A}})$ but we omit the outer terms as they are simply $\bI$ for this problem. The sample value of this quantity gives an unbiased estimate of the degrees of freedom: \[\widehat{\text{df}} = \Tr\left((\tilde{\bX}_{\mathcal{A}}^T\tilde{\bX}_{\mathcal{A}})^{-1}\right) = \Tr\left((\bI_{|\mathcal{A}|} + \alpha\bOmega_{\bu}^{\mathcal{A}})^{-1}\right).\] Rather than calculating the inverse, we substitute the first two terms of the Taylor expansion $(\bI + \bA)^{-1} = I - \bA + \bA^2 - \bA^3 + \dots$ to get 
\[\widehat{\text{df}} \approx \Tr\left(\bI_{|\mathcal{A}|} - \alpha\bOmega_{\bu}^{\mathcal{A}}\right).\]
The approximate BIC can then be obtained by substitution into the standard BIC formula \citep{Schwarz:1978,Claeskens:2008}, using the maximum likelihood estimate of the residual variance $\hat{\sigma}^2 = \frac{1}{n}\|\bX\bv - \hat{\bu}\|_2^2 = \text{RSS}/n$:
\begin{align*}
  \text{log-likelihood} &= -\frac{n}{2} \log(2\pi \sigma^2) - \sum_{i=1}^n \frac{(u_i - \bx_i^T\bv)^2}{2\sigma^2} \\
  &= -\frac{n}{2} \log(2\pi \sigma^2) - \frac{\|\bu - \bX\bv\|_2^2}{2\sigma^2} \\
  \left.\text{log-likelihood}\right|_{\hat{\sigma}^2 = \text{RSS}/n} &= -\frac{n}{2}\log(\text{RSS}/n) -\frac{n}{2}\log(2\pi) - \frac{n}{2} \\
  \left.-2 * \text{log-likelihood}\right|_{\hat{\sigma}^2 = \text{RSS}/n} &= n\log(\text{RSS}/n) + n\log(2\pi) + n \\
  \implies \text{BIC}(\hat{\bu}) &= \log\left[ \frac{1}{n}\|\bX \bv-  \hat{\bu}\|_{2}^{2}\right] + \frac{1}{n} \log(n) \,  \widehat{\text{df}}(\hat{\bu})
\end{align*}
where the $n\log(2\pi)$ and $n$ constant terms can be omitted in the BIC criterion.\qedhere

\end{proof}

\begin{figure}[h]
\centering
\includegraphics[width=\textwidth]{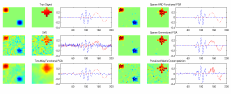}
\vspace{-0.4in}
\caption{Simulated factors used for the simulation study in Section \ref{sec:simulationII} and estimates thereof: for each sub-figure, the first two panels are the left singular vectors ($\bu_1, \bu_2$), while the third panel shows the right singular vectors, $\bv_{1}$ (red, dotted-dashed) and $\bv_{2}$ (blue, dashed). While SFPCA and SGPCA both perform well on the left singular vectors, only SFPCA is able to simultaneously identify the spatial sparsity and smooth structure of the sinusoidal pulses in the right singular vectors.}
\label{fig:simulationII}
\end{figure}
\section{Additional Results}
In this section, we extend the results given in Sections \ref{sec:simulation} and \ref{sec:eeg}.

\subsection{Two-Way Simulation Study} \label{sec:simulationII}

The simulation presented in Section \ref{sec:simulation} (Table \ref{tab:simulationI} and Figure \ref{fig:simulationI}) contain exhibit smooth and sparse structure in the right singular vectors ($\bv$), but not in the left singular vectors ($\bu$), which are selected from the unit sphere randomly (Haar measure). In this section, we demonstrate the performance of SFPCA on data exhibiting smoothness and sparstity in both $\bu$ and $\bv$ in a rank-2 model.

The true factors in this simulation are inspired by neuroimaging data with both spatial and temporal structure. $\bU \in \R^{625 \times 2}$ are the spatial factors corresponding to a $25 \times 25$ imaging grid, with $\bu_1 = \bU_{\cdot 1}$ containing two non-overlapping regions of interest with smooth edges  and $\bu_2 = \bU_{\cdot 2}$ containing a single region of interest with sharp edges. $\bV \in \R^{200 \times 2}$ are the same temporal factors used in Section \ref{sec:simulation}, namely time-localized sinusoidal pulses. These factors are shown in the top left panel of Figure \ref{fig:simulationII}. 

Data are generated as $n = 200$ samples from the low-rank model $\bX = \sum_{k=1}^2 d_i\bu_k\bv_k^T + \bE$ where the elements of $\bE$ are are independently and identically drawn from a standard normal distribution. The signal-to-noise ratio is fixed at $d_1 = n / 6$ and $d_2 = n / 7$. As before, SFPCA is compared with several competing methods, including the two-way FPCA (TWFPCA) method of \citet{Huang:2009}, the sparse SVD (SSVD) method of \citet{Lee:2010}, the penalized matrix decomposition (PMD) of \citet{Witten:2009}, and the sparse generalized PCA (SGPCA) of \citet{Allen:2014}. Each method was tuned according to the authors' recommendation, with SFPCA tuned using the greedy BIC scheme described above. For SFPCA and TWFPCA, $\bOmega_{\bu}$ is the second differences matrix over a $25 \times 25$ grid and $\bOmega_{\bv}$ is the second-differences matrix of a chain graph of length $200$ (\emph{i.e.}, a tridiagonal matrix with $(-1, 2, -1)$ on the tridiagonal). For SGPCA, the generalizing operators ($\bQ, \bR$ matrices) were again constructed from $\bOmega_{\bu}$ and $\bOmega_{\bv}$ using the methods suggested by \citet{Allen:2011}. 

Qualitative results from this study are shown in Figure \ref{fig:simulationII}, where we see that SFPCA clearly outperforms the competing methods. Results for the temporal ($\bV$) factors are similar to those for our one-way simulation, so we focus on the spatial ($\bU$) factors here. The standard SVD provides neither sparsity, nor spatial smoothness, though the outline of the true signals can be discerned. TWFPCA recovers the smooth structure spatial signals well, but is not able to provide sparsity elsewhere. PMD appears to identify the signal, but as it does not allow for spatial smoothness, is insufficiently sparse elsewhere. SFPCA and optimally tuned SGPCA (here shown with $\sigma = 1$) both perform well here, but SGPCA is unable to recover the temporal smoothness patterns in the right singular vectors. 

Quantitative results are presented in Table \ref{tab:simulationII}, where again we report the true positive rate (TP) and false positive rate (FP) for support recovery, as well as the relative angle and relative squared error to measure smoothness, which measure overall signal recovery. (See the main text for definitions). Consistent with the qualitative results, TWFPCA does well at recovering the true spatial signal in the first left singular vector, but cannot identify the sparse activation regions. Optimally-tuned SGPCA and SFPCA both perform well, with SFPCA slightly outperforming for the leading singular vectors and SGPCA outperforming for the following singular vectors. The good performance of SGPCA on this example is somewhat surprising as GPCA assumes smoothness in the noise, which is here \textsc{IID}, rather than the signal itself.

\begin{table}[ht]
\centering
\scalebox{1}{
\begin{tabular}{lr|r|r|r|r|r|r}
\toprule
\multicolumn{2}{c}{} & \multicolumn{1}{c}{TWFPCA} & \multicolumn{1}{c}{SSVD} & \multicolumn{1}{c}{PMD} & \multicolumn{1}{c}{SGPCA ($\sigma=1$)} & \multicolumn{1}{c}{SGPCA ($\sigma=5$)} & \multicolumn{1}{c}{SFPCA} \\
\midrule
\multirow{3}{*}{$\bu_1$} & TP & - & {\bf 0.944} (.004)	& 0.697 (.005) & 0.843 (.005) & 0.532
  (.004) &  0.876 (.013) \\
& FP & - & 0.611 (.111) & 0.015 (.002) & 0.024 (.002) & {\bf 0.000 } (.000)
  &   0.007 (.012) \\
& r$\angle$ & {\bf 0.0832} (.041) & 0.608 (.110) & 0.934 (.024) & 0.321 (.011) &
  1.140 (.034) & 0.356 (.084) \\
\midrule
\multirow{3}{*}{$\bv_{1}$} & TP & - & {\bf 0.852} (.004) & 0.679 (.004) & 0.629 (.005) & 0.659
  (.007) &  0.765 (.006) \\
& FP & - & 0.617 (.111) & 0.259 (.003) & {\bf 0.018} (.001) & 0.045 (.004) &
   0.055 (.033) \\
& r$\angle$ & 0.252 (.071) & 0.664 (.115) & 0.565 (.009) & 0.235 (.004) &
  0.186 (.005)  &  {\bf 0.142} (.053) \\
\midrule
\multirow{3}{*}{$\bu_{2}$} & TP & - & {\bf 0.892} (.005) & 0.751 (.006) & 0.679 (.006) & 0.031
  (.002) & 0.562 (.016) \\
& FP & - & 0.616 (.111) & 0.202 (.005) & 0.032 (.002) & {\bf 0.000} (.000) &
  0.006 (.011) \\
& r$\angle$ & 0.498 (.100) & 0.547 (.105) & 0.376 (.011) & {\bf 0.325} (.010) &
  3.650 (.088) & 0.568 (.107) \\
\midrule
\multirow{3}{*}{$\bv_{2}$} & TP & - & {\bf 0.996} (.001) & 0.983	(.003) & 0.981 (.003) & 0.659
  (.010) & 0.946 (.008) \\
& FP & - & 0.614 (.111)	& 0.256 (.002) & {\bf 0.014} (.001) & 0.058 (.005) &
  0.024 (.022) \\
& r$\angle$ & 0.720 (.120) & 0.647 (.114) & 0.439 (.007) & {\bf 0.213} (.006) &
  1.240 (.036) & 0.355 (.084) \\
\midrule
& rSE & 0.276 (.001) & 0.501 (.001) & 0.470 (.004) & {\bf 0.203} (.003) &
  0.642 (.015) & 0.212 (.001) \\
\bottomrule
\end{tabular}}
\caption{Performance of various regularized PCA methods for the simulation study described in Section \ref{sec:simulationII}. Results are averaged over 50 replicates, with standard errors given in parentheses. For each method, the true positive rate (TP), false positive rate (FP), relative angle compared to that of the SVD ($\text{r}\angle$), and relative squared error compared to that of the SVD ($\text{rSE}$) are reported. (TP and FP are not reported for the non-sparse TWFPCA.) The best performing method on each metric is bold-faced. Both SFPCA and SGPCA perform well on this example, though SFPCA has the additional advantage of not requiring the user to choose the smoothing parameter $\sigma$.} \vspace{-0.2in}
\label{tab:simulationII}
\end{table}

\subsection{Additional EEG Results}
\begin{figure}[h]
\centering
\includegraphics[width=3.25in]{fig_eeg_sfpca5.pdf}\hspace{4mm}\includegraphics[width=3.25in]{fig_eeg_ica5.pdf}\\
\includegraphics[width=3.25in]{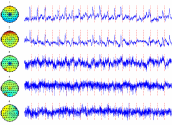}\hspace{4mm}\includegraphics[width=3.25in]{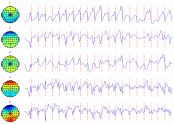}\\
\includegraphics[width=3.25in]{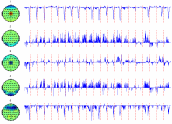}\hspace{4mm}\includegraphics[width=3.25in]{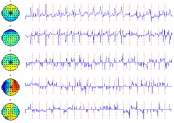}
\caption{EEG Case Study: first five spatial and temporal components as calculated by SFPCA (top-left), ICA (top-right), PCA (middle-left), TWFPCA (middle-right), TWSPCA / PMD (bottom-left), and TWSGPCA (bottom-right). The spatial and temporal signals identified by SFPCA are the most interpretable (sparsity) and the most biologically plausible (smoothness), while, unlike the other methods considered, also identifying meaningful structure past the first two components.}
\label{fig:eegII}
\end{figure}

In Section \ref{sec:eeg} of the main text, we compared the estimated SFPCA factors with ICA on electroencephalography (EEG) data from the UCI Machine Learning repository. In Figure \ref{fig:eegII}, we show the results of applying (standard) PCA, two-way FPCA \citep{Huang:2009}, two-way SPCA via the penalized matrix decomposition (TWSPCA) \citep{Witten:2009}, and two-way sparse generalized PCA (TWSGPCA) \citep{Allen:2014}. 

As noted in the main body of the text, SFPCA and ICA identify similar temporal and temporal patterns for the first two components, but the SFPCA components have superior temporal sparsity, yielding improved interpretability. Standard PCA returns similar results to ICA, again failing to identify structure after the first two components. TWFPCA identifies smooth and biologically plausible smooth signals in all components, but cannot yield sparse estimates, hindering interpretation. TWSPCA returns similar first components (recall that these estimates are only defined up to a sign factor), but returns significantly more jagged estimates for the following components. The temporal components estimated by TWSGPCA are significantly more jagged and less sparse than those returned by SFPCA and do not exhibit meaningful temporal or spatial localization. 

\clearpage
\section{Additional Background}
Since its introduction by \citet{Pearson:1901} and  \citet{Hotelling:1933}, principal component analysis (PCA) has been a mainstay of applied statistics. PCA provides a unified, computationally efficient, and mathematically elegant approach to dimension reduction, data visualization, and feature engineering. The usefulness of PCA has lead to its rediscovery by many other fields where it is variously known as the Karhunen-Lo\`eve transform \citep{Karhunen:1946,Loeve:1946} in the theory of stochastic process, the method of empirical orthogonal functions in the environmental and atmospheric sciences (at least when the observation grid is regular; see the note of \citet{Buell:1971} and the discussion thereof by \citet{Wikle:1999}), and the proper orthogonal decomposition in various engineering fields \citep{Berkooz:1993}, among other names. In its classical form, PCA is performed on a (centered) data matrix $\bX$ by taking the eigendecomposition of the scatter (covariance) matrix: $\bX^T\bX$. The Eckart-Young theorem \citep{Eckart:1936,Stewart:1993} establishes an equivalence between this formulation and the low-rank formulation we consider: \[\bX = \sum_{i=1}^k d_i\bu_i\bv_i \text{ for } \{\bu_i\}_{i=1}^k, \{\bv_i\}_{i=1}^k \text{ orthogonal elements of } \R^n, \R^p \text{ respectively.} \]
The elements of this low-rank representation can be found using the singular value decomposition of $\bX$, which can be efficiently computed for large data matrices using the algorithms of \citet{Golub:1965} and \citet{Golub:1970}, as well as many more modern variants. We favor the low-rank formulation as it captures patterns in both the rows and columns of $\bX$, but emphasize that equivalence between the two formulations holds only for the standard formulations and is broken when regularization is introduced. While we focus on matrix decomposition approaches, PCA may also be interpreted as the MLE of a certain probabilistic model, as shown by \citet{Tipping:1999}. The model-based framing is particularly useful when extending PCA to more complex data structures, \emph{e.g.}, the integrative PCA (iPCA) model for multi-block data recently proposed by \citet{Tang:2018}. 

When applying PCA, the selection of the true number of principal components is an important task. The ``scree'' heuristic of \citet{Cattell:1966} considers the rate at which the singular values of $\bX$ level off. To address the inherent subjectivity of this approach, several data-driven cross-validation-type techniques have been proposed \citep{Wold:1978,Eastment:1982,Buja:1992,Troyanskaya:2001,Owen:2009,Josse:2012}. More recently, rank selection methods based on the sampling distribution of noise eigenvalues have been proposed \cite{Rao:2008,Kritchman:2008,Choi:2017}. Several of these strategies are based on recent developments in random matrix theory which characterize the asymptotic properties of random matrices for standard null hypotheses \citep{Johnstone:2001,Paul:2014}. Before proceeding further, we note that we have only touched on a small fraction of the vast literature on PCA and refer the reader to the book of \citet{Jolliffe:2002} or the more recent review of \citet{Abdi:2010} for more a comprehensive coverage. 
The statistical properties of PCA have been studied by many authors, among which \citet{Anderson:1963} and \citet{James:1964} stand out as early and important references. These authors, and the rich theory developed afterward \citep{Muirhead:1982}, establish asymptotic consistency of PCA in the large-sample ($n \to \infty$) setting. As statistical interest in data-sets for which the ``aspect ratio'' $n / p$ is small grew, the short-comings of standard PCA were widely noted. New results in random matrix confirmed this observation: that standard PCA performs quite poorly unless the aspect ratio is large \cite{Baik:2005,Paul:2007,Johnstone:2009,Wang:2017,Dobriban:2017}. (\citet{Johnstone:2018} give a useful and accessible review of the implications of random matrix theory for PCA. \citet{Bai:2008} review the closely related literature on high-dimensional econometric factor models, among which \citet{Bai:2003} stands out as a key reference.) To address this, regularized variants of PCA were proposed, several of which were later shown to yield consistent estimates. 

The earliest forms of regularized PCA to appear in the literature arose in the functional data analysis community, where the principal components themselves were assumed to follow a smooth (functional) structure with respect to some norm. The early development of PCA of functional dates back to \citet{Dauxois:1982} and \citet{Besse:1986}, but \citet{Rice:1991} was the first, to the best of our knowledge, to explicitly impose a curvature penalty and propose an explicitly \emph{functional} PCA (FPCA). \citet{Silverman:1996} later penalized the curvature by altering the constraint region of the PCA problem and showed that this approach is equivalent to changing the norm and closely related to half-smoothing the data. From here, \citet{Huang:2008} showed that this approach can be formulated as a regression problem with a penalty on the $\bv$-terms. \citet{Huang:2009} extended this idea to the low-rank model and proposed two-way FPCA  via an alternating penalized regression scheme. They further established that this approach could be interpreted as attaining a (potentially local) solution to a penalized SVD problem. \citet{Zhang:2013} proposed a robust extension of the method of \citet{Huang:2009} where the Frobenius loss used for formulate the low-rank model is replaced by a robust loss function. \citet{Allen:2013} later extended these approaches to the multi-way (tensor) setting. The literature on FPCA is vast and we refer the reader to the books by \citet{Ramsay:2002,Ramsay:2005} and the review of \citet{Hall:2011} for more comprehensive coverage.

Sparsity-inducing regularized PCA (SPCA) was first proposed by \citet{Jolliffe:2003} who augmented the eigenvalue formulation of PCA with an $\ell_1$ (\textsc{Lasso} \citep{Tibshirani:1996}) constraint on the eigenvectors. \citet{Yuan:2013} and \citet{Ma:2013} proposed algorithmic variants of the sparse eigenvalue problem which incorporate truncation and hard-thresholding steps, respectively, into standard eigenvector algorithms. (\citet{Journee:2010} give an interesting variant of this approach which retains convexity.) Convex semi-definite relaxations of the sparse eigenvalue problem were proposed by several authors \citep{dAspremont:2007,dAspremont:2008,Vu:2013-Fantope}, while \citet{Moghaddam:2005} propose a greedy search scheme. \citet{Johnstone:2009} proposed a wavelet thresholding method which attempts to improve estimation of the covariance eigenstructure before standard PCA is performed, while \citet{Deshpande:2014} consider an algorithm based on direct covariance thresholding previously considered by \citet{Bickel:2008a,Bickel:2008b}. \citet{Wang:2014} propose an iterative approach to approximately solve the $k$-sparse eigenvalue problem with statistical guarantees. Finally, \citet{Asteris:2015} propose an intriguing method for estimating several sparse principal components based on bipartite graph matching. 

We note that, however, that many of these approaches are derived from non-convex problems, which limits their theoretical tractability and computational efficiency. Additionally, these methods require instantiating and repeated use of the sample covariance matrix, which may be expensive for large scale problems. To address this, \citet{Zou:2006-SPCA} proposed an alternative formulation which builds upon the \textsc{ElasticNet} \citep{Zou:2005} penalized regression approach but requires solving a bi-convex problem using an iterative alternating regression scheme, an computational strategy shared with the low-rank model. \citet{Gataric:2017} propose an approach based on aggregating principal components of random low-dimensional projections of $\bX$ which helps to limit the computational complexity. The majority of the methods discussed above identify only the leading principal component: if additional principal components are desired, they can be applied recursively to a ``deflated'' matrix. The most commonly used deflation method is that of Hotelling ($\bX := \bX - d\bu\bv^T$) though \citet{Mackey:2008} presents alternative approaches with better orthogonality properties. More recently, manifold optimization techniques have been used by \citet{Benidis:2016} and by \citet{Chen:2019} to simultaneously estimate multiple sparse  principal components. 

An early form of low-rank approximation with sparse factors was considered by \citet{Zhang:2002,Zhang:2004}. In the statistical literature, the low-rank model used in our SFPCA formulation was first considered by \citet{Shen:2008} for one-way sparsity and by \citet{Witten:2009} for two-way sparsity: both proposed alternating regression schemes to calculate leading singular values, though the Lagrangian form of \citet{Allen:2014} is closer to our approach. \citet{Lee:2010} and \citet{Yang:2014} proposed similar sparse singular value frameworks. \citet{Allen:2012} extended the sparse low-rank model to the multi-way (tensor) setting. \citet{Udell:2016} review a range of similar models with the squared error loss replaced by other (exponential family) losses. 

Many theoretical results for SPCA have been established in the literature, primarily for the covariance model: see, \emph{e.g.}, the papers by \citet{Amini:2009}, \citet{Jung:2009}, \cite{Cai:2013}, \citet{Vu:2012,Vu:2013}, \citet{Birnbaum:2013}, \citet{Berthet:2013}, \citet{Shen:2013}, \citet{dAspremont:2014}, \citet{Cai:2015}, \citet{Lei:2015}, \citet{Krauthgamer:2015}, \citet{Ma:2015}, \citet{Wang:2016}, and \citet{Bresler:2018}, among many others. For that reason, we do not attempt to pain a comprehensive picture here, instead referring the reader to the review paper of \citet{Zou:2018}. In addition to standard sparse PCA, several other sparse PCA variants have been proposed, including non-negative sparse PCA \citep{Zass:2006, Allen:2011}, structured-sparse PCA \citep{Jenatton:2010,Bach:2012}, sparse PCA with structured noise \citep{Allen:2014}, contamination-robust sparse PCA \citep{Croux:2013,Hubert:2016}, and distributionally-robust sparse PCA \citep{Han:2014,Han:2018}. \citet{Lu:2016} propose an extension of sparse PCA to data sampled from an exponential family, building on an early proposal of \citet{Collins:2001}. (See also the related proposals of \citet{Lee:2010-AOAS} and of \citet{Liu:2018}.) Many of these schemes are based on the \textsc{Lasso} \citep{Tibshirani:1996} penalty or structured variants thereof \citep{Yuan:2006,Jenatton:2010}, but the use of non-convex penalties has been occasionally considered: \citet{Shen:2008} compare the use of \textsc{SCAD} \citep{Fan:2001} and hard-thresholding (``$\ell_0$'') penalties in their scheme, while \citet{Lee:2012} augment the method of \citet{Zou:2006-SPCA} with a screening step followed by a penalized regression method using an \textsc{AdaptiveLasso} \citep{Zou:2006-Adaptive}, \textsc{SCAD} \citep{Fan:2001}, or \textsc{MCP} penalty \citep{Zhang:2010}. 

The combination of sparsity and smoothness that we consider has not been extensively explored in the matrix factorization literature, though \citet{Slawski:2010} and \citet{Hebiri:2011} explore similar ideas in a regression context. Based on an early draft of this paper, \citet{Li:2016} propose a framework for \emph{supervised} SFPCA which combines standard (unsupervised) SFPCA with the Supervised SVD \citep{Li:2016b} and \citet{Mohammadi:2017} propose a sparse and functional version of Canonical Correlation Analysis (CCA) \citep{Hotelling:1936}.  \citet{Chen:2015} propose a \emph{localized} FPCA which performs FPCA with the additional constraint that the estimated factors have localized support, inducing similar sparsity structures to what we observe from SFPCA. While similar in name, the multilevel sparse functional PCA of \citet{Di:2014} refers to sparsely-sampled functional data, not sparsity in the factor loadings as we consider here. 

\section{Additional References}
\printbibliography[heading=none]

\end{refsection}
\end{document}